\documentclass{article}
\newif\ifarxiv
\arxivtrue

\ifarxiv
    \usepackage{default}
\else
    \usepackage[final]{neurips_2022}
\fi

\usepackage[utf8]{inputenc} 
\usepackage[T1]{fontenc}    
\usepackage{hyperref}       
\usepackage{url}            
\usepackage{booktabs}       
\usepackage{amsfonts}       
\usepackage{nicefrac}       
\usepackage{microtype}      
\usepackage{xcolor}         
\usepackage{graphicx}
\usepackage{subcaption}
\usepackage{wrapfig}
\usepackage{xspace}

\title{Achieving and Understanding Out-of-Distribution Generalization in Systematic Reasoning in Small-Scale Transformers}

\ifarxiv
    \author{%
      Andrew J. Nam\thanks{Stanford University}, 
      Mustafa Abdool${}^{*}$,
      Trevor Maxfield${}^{*}$,
      James L. McClelland${}^{*}$
    }
\else
    \author{%
      Andrew J. Nam \\
      Department of Psychology\\
      Stanford University \\
      \texttt{ajhnam@stanford.edu} \\
      \And
      Mustafa Abdool \\
      Department of Computer Science \\
      Stanford University \\
      \texttt{moose878@gmail.com} \\
      \AND
      Trevor Maxfield \\
      Institute of Computational and Mathematical Engineering \\
      Stanford University \\
      \texttt{maxfit@stanford.edu} \\
      \And
      James L. McClelland \\
      Department of Psychology \\
      Stanford University \\
      \texttt{jlmcc@stanford.edu} \\
    }
\fi

\definecolor{green}{rgb}{0, 0.5, 0}

\def\sos{\textless{}SOS\textgreater\xspace}
\def\pad{\textless{}PAD\textgreater\xspace}
\def\eos{\textless{}EOS\textgreater\xspace}

\begin{document}

\maketitle

\begin{abstract}
Out-of-distribution generalization (OODG) is a longstanding challenge for neural networks.  This challenge is quite apparent in tasks with well-defined variables and rules, where explicit use of the rules could solve problems independently of the particular values of the variables, but networks tend to be tied to the range of values sampled in their training data. Large transformer-based language models have pushed the boundaries on how well neural networks can solve previously unseen problems, but their complexity and lack of clarity about the relevant content in their training data obfuscates how they achieve such robustness. As a step toward understanding how transformer-based systems generalize, we explore the question of OODG in small scale transformers trained with examples from a known distribution. Using a reasoning task based on the puzzle Sudoku, we show that OODG can occur on a complex problem if the training set includes examples sampled from the whole distribution of simpler component tasks.
Successful generalization depends on carefully managing positional alignment when absolute position encoding is used, but we find that suppressing sensitivity to absolute positions overcomes this limitation.  Taken together our results represent a small step toward understanding and promoting systematic generalization in transformers.
\end{abstract}

Large transformer-based `foundation' models \cite{bommasani2021opportunities} have attracted recent attention by showing some success in mathematical reasoning tasks, demonstrating a degree of systematicity and compositionality \cite{cobbe2021training, wei2022chain, lewkowycz2022solving}.
However, it is unclear how their ability to behave systematically emerges, due to the massive sizes of the training data and model parameters.  Are they demonstrating the ability to generalize out-of-distribution to novel problems?  Or are they succeeding because the data they are trained on samples from the entire space of possible training examples?

To investigate how a domain-agnostic model may learn to generalize to out-of-distribution examples, we train a small scale transformer-based network to learn solution strategies based on the popular puzzle game Sudoku.
We use a 6x6 Sudoku grid rather than the traditional 9x9, which provides sufficient complexity for investigating algorithmic reasoning while offering more tractability and lower compute requirements.
The general rule of Sudoku still applies: every $n$-celled row, column, and outlined region of the grid must contain exactly one instance of each of the $n$ alternative digits.

Sudoku is well-suited for this inquiry for several reasons.
First, it is governed by a small set of rules that are inherently abstract, relational, and form sophisticated interactions and dependencies that require careful algorithmic and deductive reasoning.
These rules form group properties and symmetries \cite{felgenhauer2006mathematics, russell2006mathematics} that translate one puzzle to another such that learning to solve a subset of examples of a class of puzzles would enable the solver to solve all puzzles with the same relational properties, provided that the abstract rules and relations are induced and the new examples can be mapped onto them.
The symmetries in Sudoku enable an elegant means to probe for systematicity by designing training and test sets that share core relational features yet differ superficially in a well-defined manner.
Second, Sudoku has been shown to be challenging to neural networks, and has only been successfully solved using a graph network architecture \cite{battaglia2018relational} with built-in domain-specific inductive biases enforcing the relevant, task-specific, symmetries \cite{palm2018recurrent}.
While this is a useful strategy for building models that solve Sudoku, it offers little guidance towards understanding how a domain-agnostic neural network can learn in a way that enables systematic generalization.
Finally, Sudoku has been used to study reasoning, learning, and generalization in humans \cite{lee2008psychological, nam2021underlies}, offering an interesting benchmark for what forms of behavior one ought to expect from a solver with human-level general intelligence.

We focus on one solution strategy in Sudoku called the Hidden Single technique and introduce a transformer neural network architecture and training set to explore out-of-distribution generalization (OODG).
Building on this network, we present the following findings:
First, a single forward pass in the network is insufficient to learn the Hidden Single strategy; sequential, multi-step reasoning is necessary.
Second, we decompose the Hidden Single strategy into two subtasks, and show that including training examples on these subtasks sampled from the full space of instances of such problems allows the model to exhibit substantial OODG of the Hidden Single strategy.
\begin{table}[bp]

\caption{Sample problems. Prompt text in black standard text. Target / model-generated text in bold. Note that rows count top to bottom and columns count left to right.}
\vspace{.1in}

\begin{minipage}{0.3\linewidth}
    \includegraphics[width=.9\linewidth]{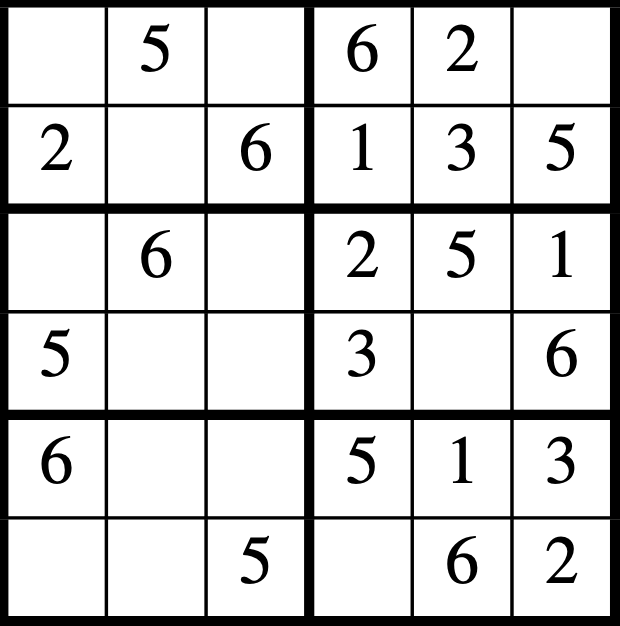}
\end{minipage}
\begin{minipage}{0.69\linewidth}
\begin{small}
\centering

\label{table:problems}
\begin{center}
\begin{tabular}{@{}ccc@{}}
\toprule
  Hidden Single &
  Full House &
  Naked Single \\ \midrule
  \begin{tabular}[t]{@{}l@{}}
    \sos hidden\_single \\
    goal\_cell row 6 column 2 \\
    house\_type column \\ 
    digit 3 \\
    can\_contain \\
    \textbf{row 1 column 2 no} \\
    \textbf{row 2 column 2 no} \\
    \textbf{row 3 column 2 no} \\
    \textbf{row 4 column 2 no} \\ 
    \textbf{row 5 column 2 no} \\ 
    \textbf{solution yes \eos}
  \end{tabular} &
  \begin{tabular}[t]{@{}l@{}}
    \sos full\_house \\
    goal\_cell row 2 column 2 \\
    house\_type box \\ 
    digit 4  \\ 
    is\_filled \\
    \textbf{row 1 column 1 no} \\
    \textbf{row 1 column 2 yes} \\ 
    \textbf{row 1 column 3 no} \\ 
    \textbf{row 2 column 1 yes} \\ 
    \textbf{row 2 column 3 yes} \\ 
    \textbf{solution no \eos}
  \end{tabular} &
  \begin{tabular}[t]{@{}l@{}}
    \sos \\
    digit 6 \\ 
    can\_contain \\
    row 4 column 3 \textbf{no} \\
    \eos
  \end{tabular} \\
  \bottomrule
\end{tabular}
\end{center}
\end{small}
\end{minipage}
\end{table}





\section{Task Description}
\label{sec:task}

We base our tasks on the three simplest Sudoku techniques. \textit{Full House (FH)} involves identifying a cell in which all other cells in one of its houses (row, column, or 2x3 box) are filled such that the empty cell's digit must be the only remaining digit. \textit{Naked Single (NS)} involves identifying a cell in which 5 of the 6 possible digits already exist in its neighborhood (row, column, and 2x3 box) such that the empty cell's digit must be the remaining digit. \textit{Hidden Single (HS)} involves identifying a cell $C$ in which all other cells $c_i$ in one of its houses cannot contain one of the digits, either due to $c_i$ already containing a different digit or the digit being present in $c_i$'s neighborhood, so that the only remaining cell that can contain the digit in the house is $C$.

For each technique, we create a task in which the model is presented with a 6x6 Sudoku grid and a string sequence prompt that provides the context for solving the problem, including the coordinates of the cell to solve for, the candidate digit, the name of the technique to use, and, for the Hidden Single (HS) and Full House (FH) tasks, the house type to inspect. By specifying all these details as part of the prompt, we simplify the task from conducting a search for a valid solution to verifying whether a goal cell should contain a candidate digit according to the rules of the specified technique.

The target sequence formats for each of the three tasks were designed to support composition of elements of the Full House and Naked Singles (NS) tasks (see Table \ref{table:problems}).
The HS task format steps through all of the cells other than the target cell in the specified target house, and checks to see if    
the specified digit can go in any of these cells.  If it cannot, the answer is yes, it must go in the target cell.
The FH task performs a similar iteration strategy, but identifies whether the cell contains a digit at all at each step.
The NS task addresses the component of the HS task not present in the FH task, which is to identify whether a given cell can contain a candidate digit based on direct contradictions within its (row, column, or box) neighborhood.  In the HS task, the network can draw on the FH strategy when it encounters full cells in the target house (column 2 in the example) and on the naked single strategy when it encounters empty cells.
\section{Experiments}
\vspace{-.1in}

We use a 3-layer transformer encoder \cite{vaswani2017attention} to which 
all grid and text embeddings are passed, and from which output vectors are then mapped to output text tokens (See \textit{Training Details} in the \textit{Supplementary Materials}).
Grid cells inputs specify the x and y coordinates of each cell and indicate if the cell is empty or if not which digit it contains.
Text tokens are pared with sinusoidal position encodings as in \cite{vaswani2017attention}.
We use teacher-forcing during training and greedy autoregressive generation during evaluation.

We first check if the network could solve the tasks by producing the final yes/no responses immediately after the prompt. After training 5 model instances on 50,000 HS puzzles uniformly sampled from the full problem space, we find that these models only solve 87.8\% of held-out puzzles, compared to the 99.9\% of models trained to produce the full sequence of reasoning steps.
Taking complete success at within-distribution generalization as a pre-requisite, we use the full sequences as exemplified in Table \ref{table:problems} in all remaining experiments, which focus on out of distribution generalization in our models.

\vspace{-0.1in}
\paragraph{Out-of-distribution generalization.}
We define within-distribution (WD) puzzles as those that conform to the restrictions used to construct the training set and out-of-distribution (OOD) puzzles as those that do not conform to these restrictions. All models included in our analyses solved held-out WD puzzles with near perfect accuracy, so we only report OODG performance in our results for conciseness.


\begin{figure}[t]
\vspace{-0.4in}
    \begin{center}
        \includegraphics[width=\linewidth]{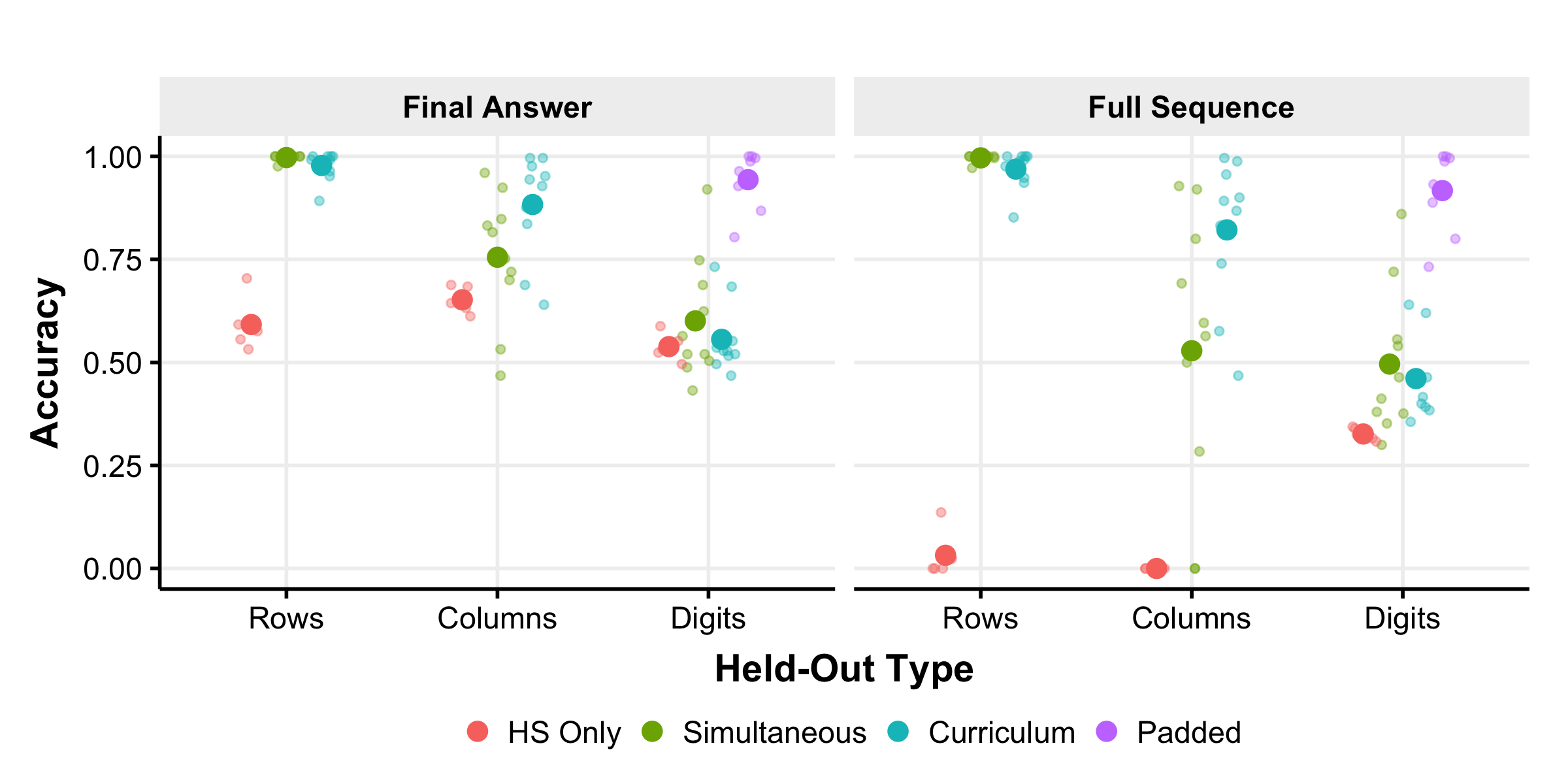}
    \end{center}
    \vspace{-0.1in}
    \caption{Out-of-distribution accuracy results. Small dots represent individual models (10 per condition). Large dots represent average accuracy in each condition. Left: accuracy based on final yes/no at the end of output sequence. Right: accuracy based on the entire output sequence.}
    \label{fig:ood-results}
    \vspace{-0.15in}
\end{figure}

\begin{wrapfigure}{r}{0.35\textwidth}
  \vspace{-0.28in}
  \begin{center}
    \includegraphics[width=0.33\textwidth]{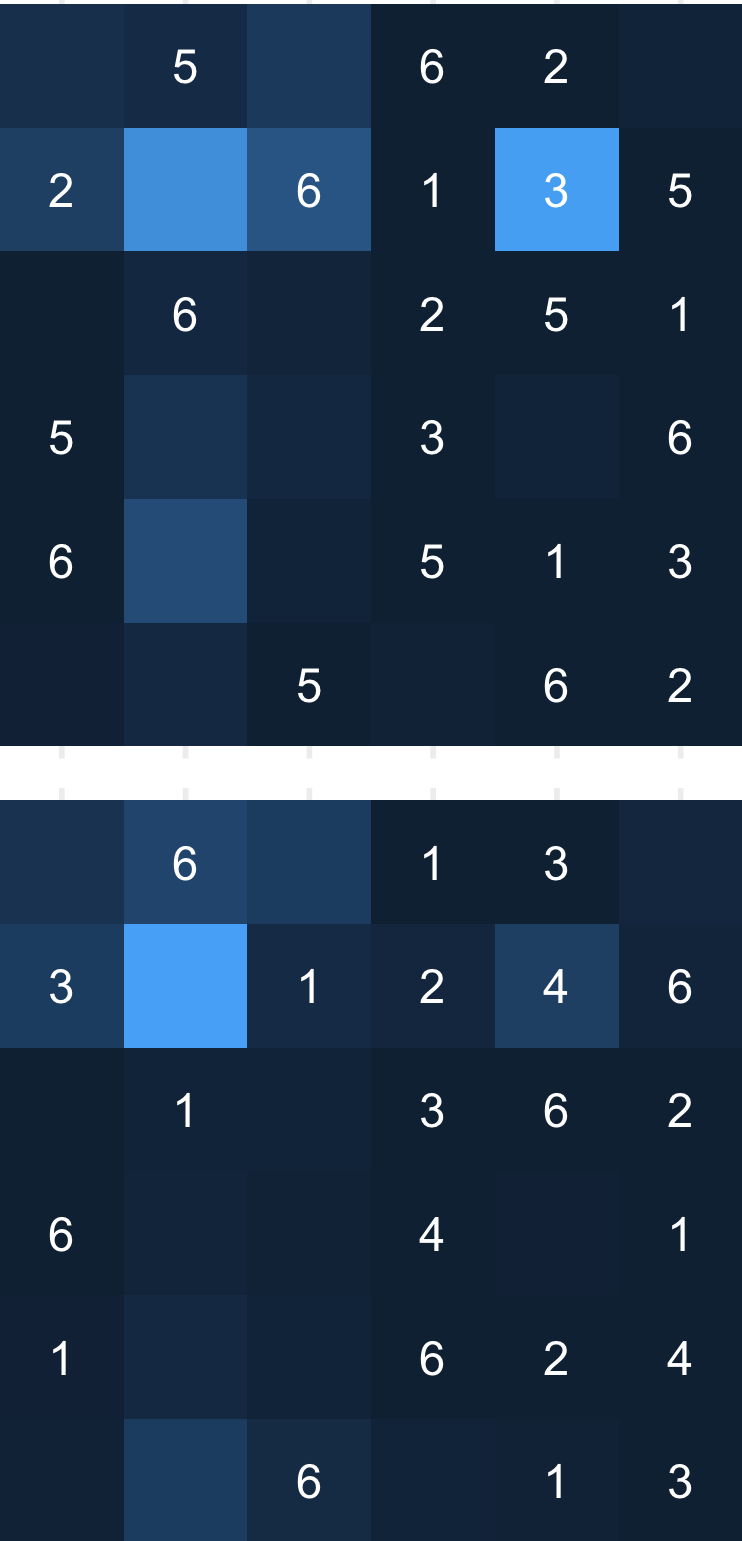}
  \end{center}
  \caption{Attentional maps. Within- and out-of-distribution puzzle with candidate digits 3 and 4 on top and bottom, respectively}
  \label{fig:attention}
  \vspace{-0.5in}
\end{wrapfigure}

Based on the natural symmetries of Sudoku grids and the train/test split used in \cite{nam2021underlies}, we devised three distributional splitting schemes to probe for OODG.
Our first condition, \textit{Rows}, recognizes that the application of the HS, FH, or NS techniques is isometric relative to the row in question.
Thus, a successful model of abstract relational reasoning should solve HS puzzles applied to any of the 6 rows, even when trained on a strict subset of the 6 rows.
We train the models using HS puzzles applied to 4 of the 6 rows (i.e. the goal cell will only appear on these 4 rows), then test its OOD performance on the remaining 2 rows.
In our second condition, \textit{Columns}, we exclude all HS puzzles that are performed over columns from the training set, and test the models on these column puzzles.  The abstract principle of process of elimination remains the same, and the main challenge is knowing which cells to iterate over.
Our final condition, \textit{Digits}, uses the fact that swapping digits only superficially changes the puzzle without affecting the underlying structure (see Figure~\ref{fig:attention}.
We train the model on puzzles with 4 of the 6 digits as the candidate digit identified in the problem prompt and test the model on puzzles with the remaining 2.

We also had 3 conditions for how we trained the models.
In the first training condition, HS Only, we included 110,000 HS puzzles as part of the training set and not the FH or NS puzzles, and trained the model for 70,000 gradient updates (each based on 192 examples).
The second condition, \textit{Simultaneous}, included 50,000 HS puzzles and 30,000 FH and NS puzzles each, and the model was trained on all 3 tasks simultaneously for 70,000 updates.
The FH and NS puzzles were sampled completely at random without any systematic constraints.
The last condition, \textit{Curriculum}, uses the same materials as Simultaneous, but trains the model on the FH and NS puzzles for the first 20,000 updates, reaching ceiling performance, before training on all 3 tasks for 50,000 more updates.

Figure \ref{fig:ood-results} summarizes the out-of-distribution generalization results in each condition.
First, we compute the accuracy based on the final yes/no answer at the end of the output sequence, and we find that models trained only on HS puzzles struggle to exceed chance (50\%), suggesting that the model has no inherent inductive bias towards generalizing in these dimensions.
When trained with the FH and NS tasks, the model succeeds in generalizing to the held-out rows and columns to a high degree, especially when trained using the curriculum-based setup.
We consider the strong generalization in the Columns condition (though somewhat less complete than in the Rows condition) to be an important finding, since the Columns condition contained no training on column-wise puzzles, while the Rows condition included 4 of the 6 rows.  The models did not generally perform well on held out digits, although one of the seeds in the simultaneous condition generalized well.  
The relational neural network \cite{palm2018recurrent} also fails to transfer to held-out digits, but humans who learn to solve HS puzzles with a restricted set of targets show no decrement in performance when tested on digits held-out from a training tutorial \cite{nam2021underlies}.

We also measure the accuracy for entire sequences, in which a problem is considered solved if the entire model output sequence matches the target sequence, including the intermediate steps.
This not only magnifies the differences, but also indicates that the models that generalize successfully do so in its entire reasoning process, not just at the correct final output.



Although the Rows and Columns conditions successfully demonstrate out-of-distribution generalization, the Digits condition generally does not, as shown by the roughly 50\% accuracy in Figure~\ref{fig:ood-results}.
This is surprising, since human participants show no change in performance when the digits in the grid are swapped \cite{nam2021underlies}.
Its peculiarity is magnified by the fact unlike the Rows and Columns conditions, full sequence accuracy is well above the floor even when the models are trained with only HS puzzles.

\paragraph{Digit generalization mis-attention.}
Inspecting the model generated outputs, we find that the models correctly identify the relevant cells to iterate over, and the last line indicating the final answer is consistent with its intermediary outputs.
In other words, the errors are made when determining whether or not the cells can contain the candidate digit. Moreover, these errors only occur at empty cells, not at cells already containing a digit.

To gain further insight, we analyze the attention maps generated by the transformer that indicate which information the model considers.
We probe the model by taking held-out puzzles and rotating the digits in each grid such that 1 becomes 2, 2 becomes 3, 6 becomes 1, etc., thus producing 6 identical sets of puzzles that only differ by the digits involved, and each puzzle in the set has a unique candidate digit.
We inspect how the model queries the grid as it produces the yes/no responses at the end of each substep in the HS problem by taking the maximum attention weight given for each cell from the last transformer layer.
Figure~\ref{fig:attention} shows an example of one model's attentional map on the HS problem shown in Table~\ref{table:problems} when considering whether the candidate digit can be placed at the cell on (2, 2).
In the within-distribution problem, where 3 is the candidate digit, the model attends to and correctly identifies the 3 in the same row.  In contrast, the model does not attend to the same position in the out-of-distribution puzzle where 4 is the candidate digit. This form of mis-attention is characteristic of the errors in the Digits condition.

\begin{table}[ht]

\caption{Aligned Naked Single problems. Hidden Single problem included for reference. Prompt text in standard text. Target / model-generated text in bold.}
\vspace{-.2in}

\label{table:naked_singles}
\begin{center}
\begin{small}
\begin{tabular}{@{}ccc@{}}
\toprule
  Hidden Single & Naked Single Unpadded & Naked Single Padded \\
\midrule
  \begin{tabular}[t]{@{}l@{}}
    \sos hidden\_single\\ goal\_cell row 6 column 2\\ house\_type column\\ digit 3\\ can\_contain \\
    \textbf{row 1 column 2 no} \\
    \textbf{row 2 column 2 no} \\
    \textbf{row 3 column 2 no} \\
    \textbf{row 4 column 2 no} \\
    \textbf{row 5 column 2 no} \\
    \textbf{solution yes \eos}
  \end{tabular} &
  \begin{tabular}[t]{@{}l@{}}
    \sos \\ digit 6\\ can\_contain\\ 
    row 4 column 3 \textbf{no} \\
    row 2 column 2 \textbf{no} \\
    row 5 column 3 \textbf{yes} \\
    row 4 column 5 \textbf{no} \\
    row 4 column 2 \textbf{yes} \\
    \eos
  \end{tabular} &
  \begin{tabular}[t]{@{}l@{}}
    \sos \pad \\ \pad \pad \pad \pad \pad \\ \pad \pad \\ digit 3\\ can\_contain \\
    row 4 column 3 \textbf{no} \\
    row 2 column 2 \textbf{no} \\
    row 5 column 3 \textbf{yes} \\
    row 4 column 5 \textbf{no} \\
    row 4 column 2 \textbf{yes} \\
    \eos
  \end{tabular} \\
\bottomrule
\end{tabular}
\end{small}
\end{center}
\vspace{-.2in}
\end{table}

\paragraph{Improving digit generalization.}
The model apparently fails to transfer its success in the NS task to correctly perform the NS task when it is embedded within the HS task.  We trace the source of this difficulty to a superficial difference between the prompt sequences used in the NS and HS tasks.  Unlike the FH task which is aligned token-for-token with the HS task, the NS task requires much fewer steps and has a shorter prompt compared to the HS and FH tasks.  Furthermore, the position of the digit in the NS prompt is not aligned with its position in the FH and HS prompts.  We test the importance of this by padding the NS prompt with null tokens so that the position of the ``digit'' token is aligned with the position in the HS and FH prompts.  We also added 4 extra ``row $r$ column $c$ yes/no'' lines in the target sequence with random coordinates to match the format of the other two tasks. Training the models on all 3 tasks simultaneously on this \textit{Padded} dataset allows them to reach 94.4\% final answer accuracy and 91.7\% full sequence accuracy in the Digits condition.  We also tried adding the 4 extra row-column query lines but not the null-token padding does not help the model generalize at all, and found that this leaves the OODG accuracy near 50\%. This suggests that the knowledge transfer from the NS to the HS tasks is impeded by the model's over-reliance on the token position as encoded with the sinusoidal encoding scheme.  It appears that the model learns to attend to the digit aligned with a specific position token in the FH task, and transfers this reliance to the HS task; while in the NS task (unless padding is introduced to align the tokens) the model relies on a different position token, and has not learned to read the held-out digits from this position.  

\begin{wrapfigure}{r}{0.5\textwidth}
  \vspace{-0.35in}
  \begin{center}
    \includegraphics[width=0.5\textwidth]{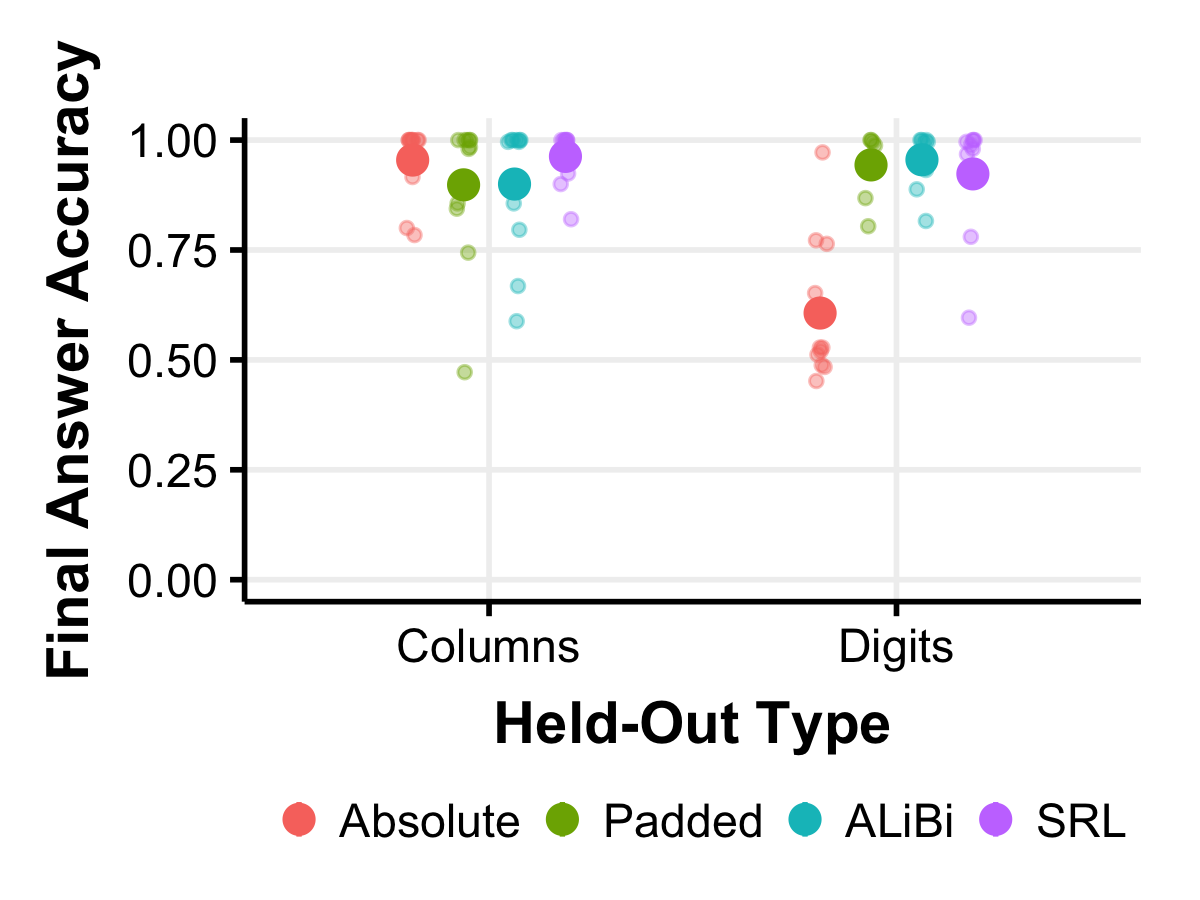}
  \end{center}
  \vspace{-0.25in}
  \caption{Out-of-distribution accuracy results with various positional encoding methods. Absolute, ALiBi, and SRL methods all train on the unpadded dataset.}
  \label{fig:pos_encoding}
  \vspace{-0.1in}
\end{wrapfigure}

However effective padding is in our Sudoku tasks, it is an impractical general solution for improving alignment between tasks for transfer learning. To address this issue, we test alternative approaches to handling positional information that have been found to reduce sensitivity to absolute token positions. Because causal masking allows transformers to learn absolute positions of tokens even in the absence of any positional information in the inputs \cite{haviv2022transformer}, we selected methods that not only remove absolute position information, and that may bias the model against relying on these position tokens anyway.
Specifically, we use sorted random labels (SRL) \cite{li2022systematic}, which replace exact position indices with an ordered sequence of position labels sampled from a larger set, and attention with linear biases (ALiBi) \cite{press2021train}, which eliminates position vectors altogether in favor of adding biases to attention scores that scale with relative distance between tokens. We apply these methods to the unpadded dataset to test whether these enable out-of-distribution generalization to held-out digits and find that final answer accuracy increases to 95.5\% and 92.3\% for ALiBi and SRL respectively.

\section{Discussion}
\vspace{-.05in}
We find, using carefully designed datasets utilizing isomorphic symmetries in Sudoku, that our transformer-based model generalized well to out-of-distribution Hidden Singles puzzles when trained in a concurrent training regime including exposure to the full distribution of the Full House and Naked Single subtasks.  These results may help shed light on the generalization abilities of larger transformer models, where it is hard to know what is covered in their training set, but where it seems reasonable to believe that exposure to component tasks and complex compositions of component tasks are interspersed throughout the data.  
While these models may receive restricted experience with complex multi-step problems, their ability to solve new ones may depend in part on more complete coverage of component sub-problems in their training data.  
Future research should explore this hypothesis in larger and more naturalistic data sets.

The initial version of our model used positional input tokens tied to absolute input position, and we found that this led to undue reliance on the exact input position of the candidate digit the model was asked to solve for, impeding generalization.    By using alternative approaches to position encoding, we were able to overcome this problem.   The work contributes to the growing body of findings indicating that failures of generalization in algorithmic tasks can be due at least in part to features of the positional encoding scheme used.  This issue is less likely to arise in a large transformer predicting long strings of tokens using variants of similar problems that would vary naturally in string length and therefor prevent reliance on exact input positions.   Nevertheless, 
these findings may be relevant to understanding the performance of large transformer-based models, especially in large language models such as GPT-3 \cite{brown2020language} where suppressing sensitivity to token positions seems to improve perplexity scores \cite{haviv2022transformer}.  Future research that corrects deficiencies related to positional encoding could have important implications for the success of larger models, perhaps helping them to achieve stronger performance in abstract reasoning than they have achieved thus far.


In any case, we are still a long way from understanding when and whether transformers are truly capable of few-shot learning and a high level of out-of-distribution transfer of the kind we have observed in some human participants in the hidden single task \cite{nam2021underlies}.  While GPT-3 was initially presented as a successful few shot-learner \cite{brown2020language}, many instances of the tasks the model supposedly learned few shot might have been embedded in its training data, and prompting with few shot examples might better be seen as a way of priming the model to be more likely to generate sequences corresponding to a task it has been exposed to in its training data many times.  Even when such models are fine tuned with instructions and few-shot examples there is considerable uncertainty about exactly why such fine tuning is helpful \cite{min2022rethinking,webson2021prompt}, and much more work is needed to clarify these issues.

\newpage
\ifarxiv
    \bibliographystyle{plain}
    \bibliography{main}

\begin{thebibliography}{10}

\bibitem{battaglia2018relational}
Peter~W Battaglia, Jessica~B Hamrick, Victor Bapst, Alvaro Sanchez-Gonzalez,
  Vinicius Zambaldi, Mateusz Malinowski, Andrea Tacchetti, David Raposo, Adam
  Santoro, Ryan Faulkner, et~al.
\newblock Relational inductive biases, deep learning, and graph networks.
\newblock {\em arXiv preprint arXiv:1806.01261}, 2018.

\bibitem{bommasani2021opportunities}
Rishi Bommasani, Drew~A Hudson, Ehsan Adeli, Russ Altman, Simran Arora, Sydney
  von Arx, Michael~S Bernstein, Jeannette Bohg, Antoine Bosselut, Emma
  Brunskill, et~al.
\newblock On the opportunities and risks of foundation models.
\newblock {\em arXiv preprint arXiv:2108.07258}, 2021.

\bibitem{brown2020language}
Tom~B. Brown, Benjamin Mann, Nick Ryder, Melanie Subbiah, Jared Kaplan,
  Prafulla Dhariwal, Arvind Neelakantan, Pranav Shyam, Girish Sastry, Amanda
  Askell, Sandhini Agarwal, Ariel Herbert{-}Voss, Gretchen Krueger, Tom
  Henighan, Rewon Child, Aditya Ramesh, Daniel~M. Ziegler, Jeffrey Wu, Clemens
  Winter, Christopher Hesse, Mark Chen, Eric Sigler, Mateusz Litwin, Scott
  Gray, Benjamin Chess, Jack Clark, Christopher Berner, Sam McCandlish, Alec
  Radford, Ilya Sutskever, and Dario Amodei.
\newblock Language models are few-shot learners.
\newblock In {\em Advances in Neural Information Processing Systems 33: Annual
  Conference on Neural Information Processing Systems 2020, NeurIPS 2020,
  December 6-12, 2020, virtual}, 2020.

\bibitem{cobbe2021training}
Karl Cobbe, Vineet Kosaraju, Mohammad Bavarian, Jacob Hilton, Reiichiro Nakano,
  Christopher Hesse, and John Schulman.
\newblock Training verifiers to solve math word problems.
\newblock {\em CoRR}, abs/2110.14168, 2021.

\bibitem{felgenhauer2006mathematics}
Bertram Felgenhauer and Frazer Jarvis.
\newblock Mathematics of sudoku i.
\newblock {\em Mathematical Spectrum}, 39(1):15--22, 2006.

\bibitem{haviv2022transformer}
Adi Haviv, Ori Ram, Ofir Press, Peter Izsak, and Omer Levy.
\newblock Transformer language models without positional encodings still learn
  positional information.
\newblock {\em arXiv preprint arXiv:2203.16634}, 2022.

\bibitem{kingma2014adam}
Diederik~P Kingma and Jimmy Ba.
\newblock Adam: A method for stochastic optimization.
\newblock {\em arXiv preprint arXiv:1412.6980}, 2014.

\bibitem{lee2008psychological}
NY~Louis Lee, Geoffrey~P Goodwin, and Philip~N Johnson-Laird.
\newblock The psychological puzzle of sudoku.
\newblock {\em Thinking \& Reasoning}, 14(4):342--364, 2008.

\bibitem{lewkowycz2022solving}
Aitor Lewkowycz, Anders Andreassen, David Dohan, Ethan Dyer, Henryk
  Michalewski, Vinay Ramasesh, Ambrose Slone, Cem Anil, Imanol Schlag, Theo
  Gutman-Solo, et~al.
\newblock Solving quantitative reasoning problems with language models.
\newblock {\em arXiv preprint arXiv:2206.14858}, 2022.

\bibitem{li2022systematic}
Yuxuan Li and James~L McClelland.
\newblock Systematic generalization and emergent structures in transformers
  trained on structured tasks.
\newblock {\em arXiv preprint arXiv:2210.00400}, 2022.

\bibitem{min2022rethinking}
Sewon Min, Xinxi Lyu, Ari Holtzman, Mikel Artetxe, Mike Lewis, Hannaneh
  Hajishirzi, and Luke Zettlemoyer.
\newblock Rethinking the role of demonstrations: What makes in-context learning
  work?
\newblock {\em arXiv preprint arXiv:2202.12837}, 2022.

\bibitem{nam2021underlies}
Andrew~Joohun Nam and James~L McClelland.
\newblock What underlies rapid learning and systematic generalization in
  humans.
\newblock {\em arXiv preprint arXiv:2107.06994}, 2021.

\bibitem{palm2018recurrent}
Rasmus Palm, Ulrich Paquet, and Ole Winther.
\newblock Recurrent relational networks.
\newblock {\em Advances in neural information processing systems}, 31, 2018.

\bibitem{press2021train}
Ofir Press, Noah~A Smith, and Mike Lewis.
\newblock Train short, test long: Attention with linear biases enables input
  length extrapolation.
\newblock {\em arXiv preprint arXiv:2108.12409}, 2021.

\bibitem{russell2006mathematics}
Ed~Russell and Frazer Jarvis.
\newblock Mathematics of sudoku ii.
\newblock {\em Mathematical Spectrum}, 39(2):54--58, 2006.

\bibitem{vaswani2017attention}
Ashish Vaswani, Noam Shazeer, Niki Parmar, Jakob Uszkoreit, Llion Jones,
  Aidan~N Gomez, {\L}ukasz Kaiser, and Illia Polosukhin.
\newblock Attention is all you need.
\newblock {\em Advances in neural information processing systems}, 30, 2017.

\bibitem{webson2021prompt}
Albert Webson and Ellie Pavlick.
\newblock Do prompt-based models really understand the meaning of their
  prompts?
\newblock {\em arXiv preprint arXiv:2109.01247}, 2021.

\bibitem{wei2022chain}
Jason Wei, Xuezhi Wang, Dale Schuurmans, Maarten Bosma, Ed~Chi, Quoc Le, and
  Denny Zhou.
\newblock Chain of thought prompting elicits reasoning in large language
  models.
\newblock {\em arXiv preprint arXiv:2201.11903}, 2022.

\end{thebibliography}
\else
    \bibliographystyle{unsrt}
\fi

\newpage
\section{Supplementary Materials}

\subsection{Training Details}
\label{sec:sup:training}
The core of our model is a 3-layer transformer encoder \cite{vaswani2017attention} supported by embedding layers for grid digits, grid cell positions, and input text, and finally an output text decoder layer.  The input for each cell consists of three one hot vectors: six-unit vectors for the row and column, and a seven-unit vector for the digit in the cell, with the seventh used when the cell is empty.   
The row and column coordinates are encoded by a 128-dim embedding layer and the resulting vectors are concatenated as a single 256-dim vector.  The digits are encoded with a 256-dim embedding layer and the digit and coordinate vectors are summed to form a single 256-dim grid cell embedding.
All text tokens are encoded by a 256-dim embedding layer and position information is added to the vectors using the sinusoidal positional encoding scheme introduced in \cite{vaswani2017attention}.
The 36 grid cell vectors and all token vectors are passed to the transformer, which is composed of 3 encoder layers with 8 heads and 1024-dim feed-forward layers, and the output vectors are decoded using a linear layer to form the final logit values for the predicted output tokens.

During training, we use teacher-forcing to predict the next token at each sequence position and cross-entropy with the target sequence.
We mask the loss so that in the loss is only applied after the 'digit' token in the hidden single and full house tasks, and after the column number token in the naked single tasks.
The loss is computed using cross-entropy and the model is optimized using Adam \cite{kingma2014adam} with a learning rate of 0.0001. When training with multiple tasks at once, we sum the losses in one batch from each of the tasks before computing the gradient for backpropagation.  We keep the same batch size of 192 samples for each task, regardless of how many tasks are trained at once.


\subsection{Attention Map}
To obtain the attention map as shown in Figure~\ref{fig:attention}, we evaluated the model on the Hidden Single problem shown in Table~\ref{table:problems}.
We generated 5 additional problems based on the problem in Table~\ref{table:problems} by shifting the digit 1 to 5 times such that what was originally a 1 would be a 2 in the second problem, then a 3, and so on, yielding 6 puzzles that are identical in every way except the individual digits involved.
For example, the top figure in Figure~\ref{fig:attention} shows the problem as is shown in Table~\ref{table:problems}, whereas the bottom figure shows the same puzzle with all digits incremented by 1, wrapping around 6 back to 1.
In the bottom puzzle, the prompt would state ``digit 4'' to account for the increment.

After evaluating the model, we take the attentional weights from the final transformer layer where the input token was the second ``2'' from the line stating ``row 2 column 2'', since the output of this position would be a ``yes'' or a ``no''. To visualize the attention across all 8 heads, we take the maximum attention weight so that if a single head attended highly to the position, it would appear in the figure. Figure~\ref{fig:attention} only shows attention to the 36 cells in the grid for visualization purposes, though the model could and does attend to other tokens in the prompt and output sequence.

\end{document}